\documentclass[a4paper, 10pt, conference]{ieeeconf}      

\IEEEoverridecommandlockouts                              

\usepackage{soul}
\usepackage{graphicx} 
\usepackage{tikz}
\usepackage{subcaption}
\usepackage{fancyhdr}
\usepackage{multicol}
\usepackage{enumerate}
\usepackage{ragged2e}
\usepackage{titlesec}
\usepackage[]{xcolor}

\usepackage{hyperref}
\usepackage{afterpage}
\usepackage{float}
\title{\LARGE \bf
The Effect of Robot Posture and Idle Motion on Spontaneous Emotional Contagion during Robot-Human Interactions}

\author{Isabel Casso$^{1}$, Bing Li$^{1}$, Tatjana Nazir$^{1}$, Yvonne N. Delevoye-Turrell$^{1}$
 \thanks{$^{1}$Univ. Lille, CNRS, UMR 9193 - SCALab - Sciences Cognitives et Sciences Affectives,
F-59000 Lille, France}%
}

\begin{document}

\maketitle
\thispagestyle{empty}
\pagestyle{empty}

\begin{abstract}

In the next decade, social robots will be implemented in many public spaces to provide services to humans. 
We question the properties of these social robots to afford acceptance and spontaneous emotional interactions.
More specifically, in the present study, we report the effects of idle motion frequency in a robot on emotional contagion, in a face-to-face interactive task with a human participant.

The robotic system Buddy was programmed to adopt a sad posture and facial expression while telling three sad stories and moving its head up/down at low, medium, and high frequency. 
Each participant (N=15 total) was invited to sit in front of Buddy and listen to the stories. Unconscious changes in posture in the human participant were recorded using a 3D motion capture system (Qualysis).
Results show greater inclinations of the shoulder/torso towards the ground in low-frequency trials and more rigid postures in high-frequency trials. The quantity of spontaneous movement was also greater when Buddy moved at slow frequencies. These findings echo results reported in experimental psychology when two individuals are engaged in social interactions.
The scores obtained in the Godspeed questionnaire further suggest that emotional contagion may occur when Buddy moves slowly because the robotic system is perceived as more natural and knowledgeable, e.g., at a speed coherent with the expressed emotion.
Our work opens the question of the degree of importance of body posture and frequency of idle motion in the conception of robotic systems. Such additions could provide social robots that afford emotional contagion in effortless robot-human collaborative tasks.

\end{abstract}
\section{INTRODUCTION}
Humans tend to engage in spontaneous unconscious movements during social human-human interactions. When we are on the receiver side of the interaction, we tend to nod our head or mimic the changes in postures of the speaker. These gross behaviors are social indicators that we are engaging attention in the exchange. However, there are also micro changes in body sway, with slow shifts in body postures at frequencies that echo the emotional ambiance of the social exchange. Such idle motions have been defined as active states of communication \cite{egges2004} and can offer effortless collaborative situations.

Idle motions can influence the perception of a person's emotional state and thus, modulate acceptance in social situations. They are related to the intensity of energy of the speaker with frequencies that are greater in high-energy emotions (anger and joy) than in low-energy emotions (sadness and contentment). Applying such idle movements to artificial agents may influence acceptance as human individuals perceive the robotic system as more animated and alive \cite{kocon2020head}\cite{asselborn2017}\cite{vollmer2015alignment}. 

In the present study, we wanted to test whether manipulating the frequency of the idle motion would modify how human participants perceived a robot that was telling a sad story about its past life experiences. We controlled the frequency of the idle motion and measured the spontaneous body oscillations and shifts in body posture of the human participant while she/he was attentive to the negative content of the stories.  
We hypothesized that more emotion contagion would be observed in human participants when the robot oscillated at a low frequency as the expressed state of emotion was sad (low energy). We also hypothesized that a frizzing state of emotion, i.e., inhibition of spontaneous oscillations, should be observed in the human participants when the robot moved at high idle frequencies because such situations should be perceived as a threat from an angry robot (high energy). A total of three stories were used to give time for the emotional contagion to take place.

\begin{figure}[H]
    \centering
    \includegraphics[scale=0.35]{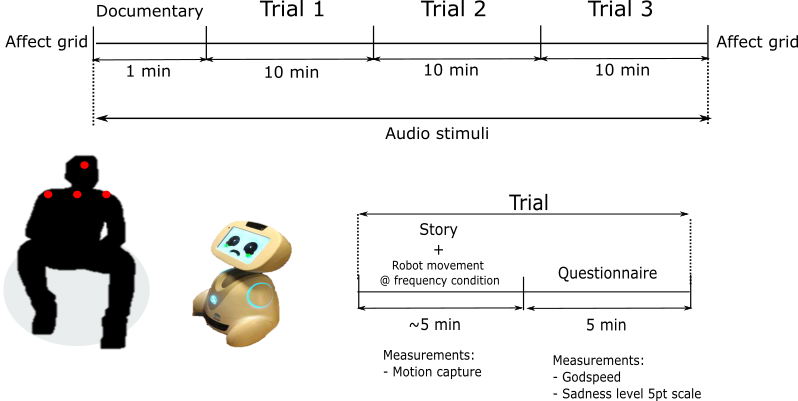}
    \caption{\small \textit{Experimental paradigm and positioning of motion capture markers on a participant}}
    \label{fig:exp}
\end{figure}

\section{METHODS}
\subsection{Robotic system}
Buddy is a social robot developed by Blue Fog robotics capable of displaying facial expressions, generating speech, and moving independently across environments.

\subsection{Experimental paradigm}
As depicted in Figure \ref{fig:exp}, the participant was invited to sit on an ergonomic ball and watch a fictional 1-minute video documentary about the life of Buddy, the robot.
Each participant was then instructed to listen to three negative stories told by the robot in the first person, with sad music playing in the background \cite{vieillard2008}. At the end of each story, participants were invited to reply to a short series of questionnaires to self-evaluate their 
experience. 

This study followed the ethical principles of the WMA Declaration of Helsinki for research involving humans. Each participant signed a consent form before inclusion and was debriefed about the aims of the study. The unique session lasted a total of 40 minutes.

While narrating the stories, Buddy moved its head up/down in the sagittal plane at three different frequencies: low (16s period), medium (12s period), and high (4s period). The idle frequencies were counterbalanced across the participants while the stories were told in the same pre-set order. A total of 15 participants were tested individually. They were college French-native students with little experience with robot systems and were aged 22-26 years.
\subsection{Material and measures}

The following measures were taken to assess the degree of emotional contagion:

\begin{itemize}
    \item \textbf{Physiological measures}
Five infrared motion capture cameras were used to collect the 3D coordinates of four reflective markers placed on each participant's shoulders, the base of the neck, and the forehead. We used the Qualisys Track Manager (QTM) software to calibrate, record, and measure body movements. 
    \item \textbf{Questionnaires}
Before and after the experiment, the participant was invited to report the valence and arousal level of their
affective states using the affect grid derived from the circumplex model of affect \cite{russell1980}.

At the end of each trial, the participant indicated the sadness level of the narration on a 0 to 5-point scale. The GODSPEED questionnaire was also filled out \cite{bartneck2009}, which offered the means to evaluate the perception of artificial agents within five characteristics: anthropomorphism, likeability, animacy, perceived intelligence, and perceived safety.
Each category comprised five semantic differential scales (e.g., Fake-Natural, Apathetic-Responsive, Unfriendly-Friendly).

\end{itemize}

\section{Movement data analysis and results}
The motion capture data was recorded at a sampling frequency of 200 Hz.
Marker velocity was extracted to compute 3D acceleration vectors. 
A filtering procedure was applied to extract all absolute acceleration peaks above 1000 $mm/s^2$; the remaining data series were interpolated before further analyses.

We computed the torso inclination
as a function of idle frequency (high; medium; low) for each trial and participant using the 
shoulder markers. 

\begin{figure}[H]
    \centering
    \includegraphics[scale=0.4]{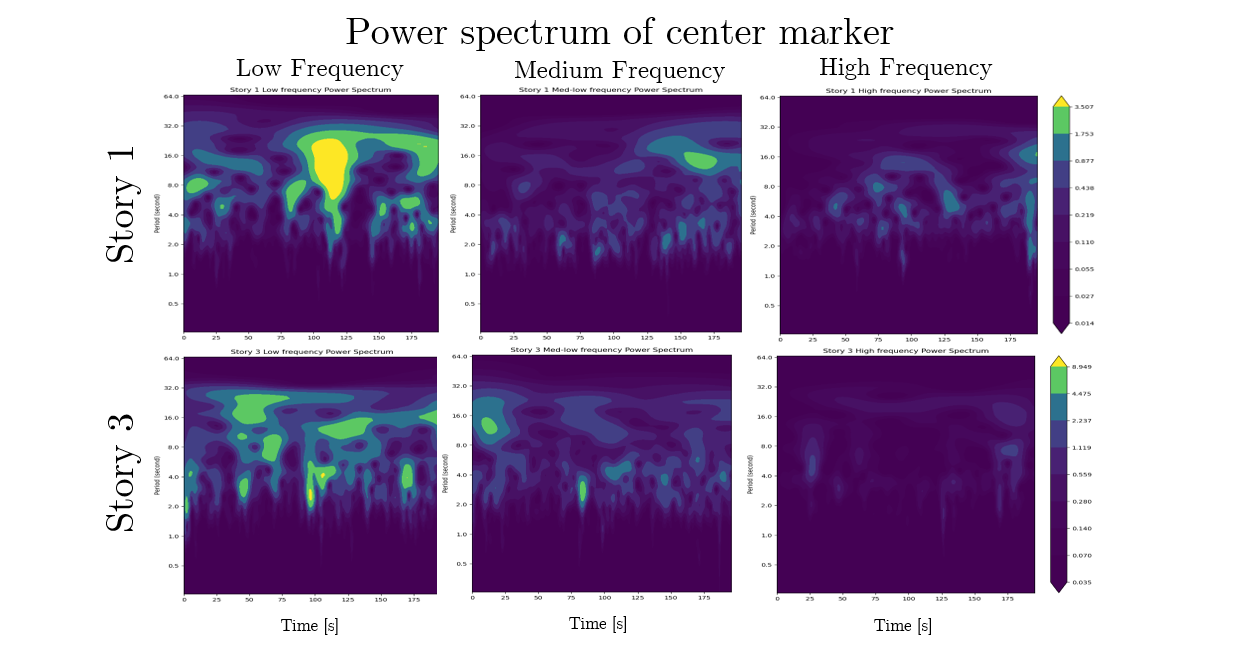}
    \caption{\small \textit{Power spectrum density of center marker averaged across participants for each trial}}
    \label{fig:ps}
\end{figure}

A time-series spectral analysis was then calculated by applying wavelet transforms on the chest marker to extract body motion. 
As shown in Figure \ref{fig:ps}, we grouped the graphs of the power spectrum of each story (presented in rows) as a function of idle frequency (presented in columns), averaged across participants.
The y-axis represents the period of the center marker in seconds as a function of the trial time, for the total story duration, in seconds. The average power spectral density (PSD) is represented with a color bar with brighter zones illustrating the highest power. 

Results showed that participants looked more to the ground at the end than at the start of a trial (M=1.8 deg). 

Participants did not engage in visible spontaneous 
torso
oscillations with Buddy (see the absence of frequencies at 4, 8, and 12s in Figure \ref{fig:ps}). Nevertheless, the spectral analysis revealed a greater power density in the low idle conditions and an absence of movement in the high idle condition. This pattern of results was observed across all three stories, even if only stories 1 and 3 are presented in Figure \ref{fig:ps}.
These findings suggest that participants tended to relax and engage in spontaneous oscillations when Buddy moved slowly; when idle was at high frequency, participants tended to freeze and inhibit spontaneous oscillations. 

\begin{figure}[H]
    \centering
    \includegraphics[scale=0.45]{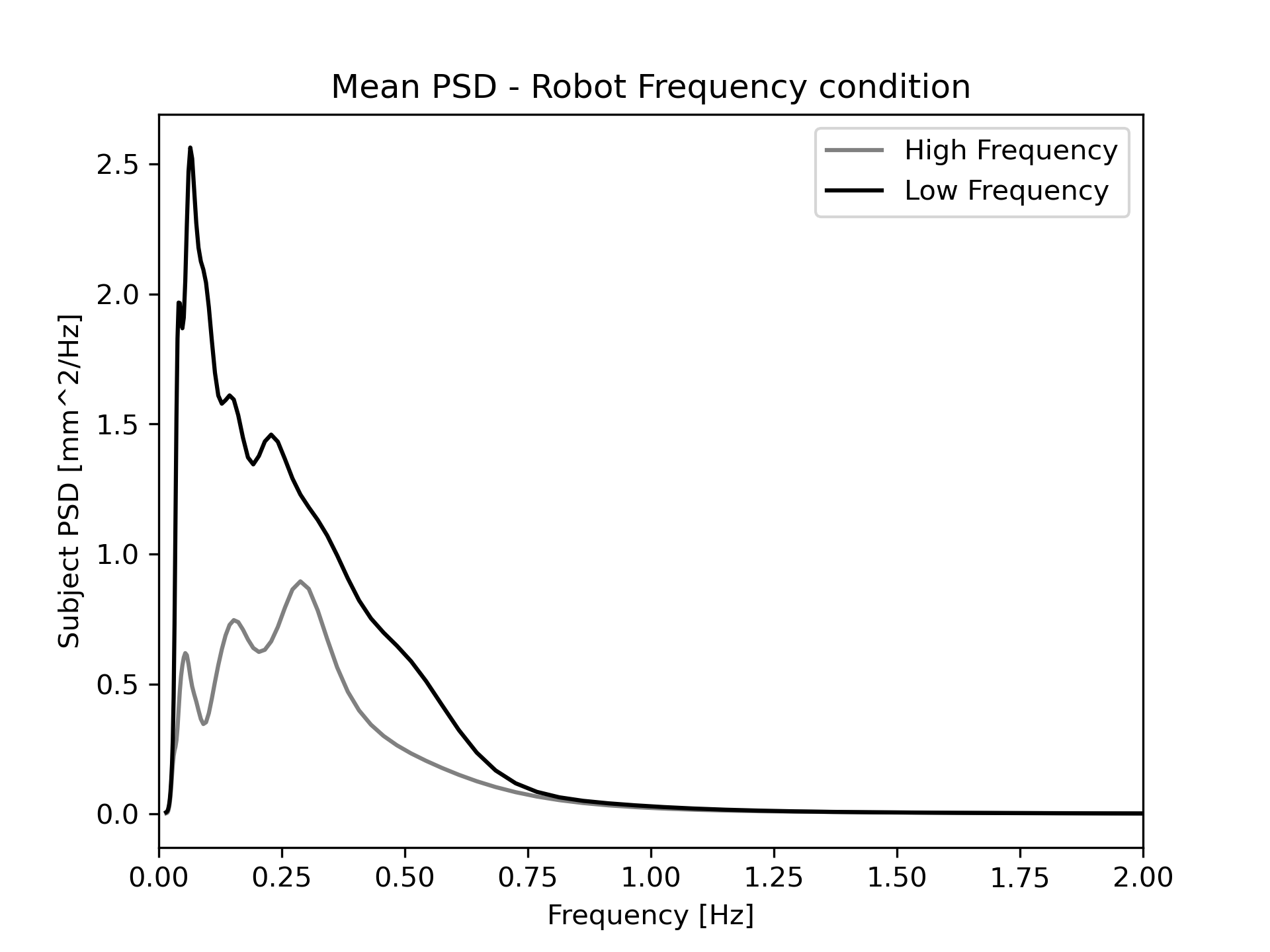}
    \caption{\small \textit{Mean power spectral density as a function of period interval for high (a) and low (b) idle frequency of the robot}}
    \label{fig:psd_curve}
\end{figure}

Figure \ref{fig:psd_curve} confirms that at high idle frequency, participants moved much less than for trials at low idle frequency. Nevertheless, in the high idle condition, movements at high frequency were more frequent than those at low frequency, confirming the presence of emotional contagion, even if to a lesser degree.    
This phenomenon resembles the posture changes reported when humans experience sadness or angry states during human-human interactions \cite{barsade2018emotional}. 

\section{Questionnaire analysis and results}

\begin{figure}[H]
    \centering
    \subfloat[\label{fig:anth}]{\includegraphics[scale=0.5]{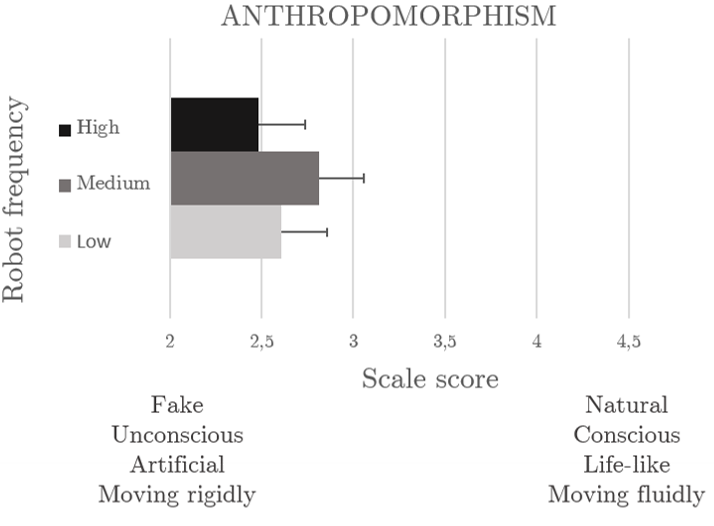}}
    \qquad
    \subfloat[\label{fig:pi}]{\includegraphics[scale=0.5]{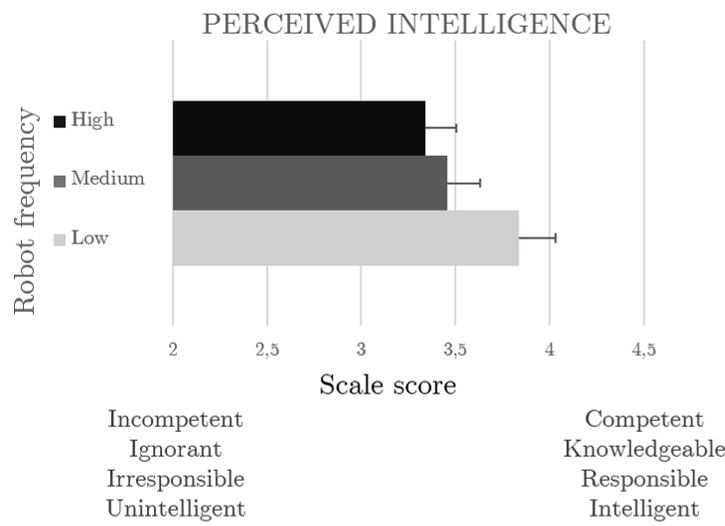}}
    \caption{\small \textit{Average scores for anthropomorphism and perceived intelligence of Buddy}}
    \label{fig:categories}
\end{figure}

The Perceived Safety category of the GODSPEED was taken out as Buddy is small and harmless in appearance. For brevity, we only report the scores that obtained significant statistical differences, i.e., the Anthropomorphism and the Perceived intelligence categories (see Figure \ref{fig:anth}). Results are shown as a function of idle frequency (y-axis) and the mean score (x-axis). 

An ANOVA on the GOODSPEED individual scores revealed a main idle effect for Anthropomorphism $(p<0.001)$ and Perceived Intelligence $(p<0.001)$ \cite{asselborn2017}. Participants perceived Buddy to be more anthropomorphic and more intelligent when it moved slowly (see Figure 4a). The analysis showed no effects of idle frequency on Likeability and Animacy. Such findings corroborate results reported previously in studies assessing emotional alignment during human-robot interactions \cite{vollmer2015alignment}.  


\section{DISCUSSION}
In the not-so-far future, Robots will become more present in our environment. However, we must scrutinize how the interaction with these systems will influence us, humans \cite{michalowski2007dancing}. 
In this study, we questioned whether the oscillatory movement of a robot's head could modulate the spontaneous engagement of a human during face-to-face interaction. More specifically, we report how the idle motion of Buddy's head can modulate emotional contagion and thus, impact a person's affective state. 
In addition, our results suggest that Buddy is perceived as more anthropomorphous and intelligent when moving slowly. Future studies will need to test whether such findings are also found when using positive induction methods during human-robot interactions.


\bibliographystyle{ieeetr}
\bibliography{Bibliography}

\begin{thebibliography}{1}

\bibitem{egges2004}
A.~Egges, T.~Molet, and N.~Magnenat-Thalmann, ``Personalised real-time idle
  motion synthesis,'' in {\em 12th Pacific Conference on Computer Graphics and
  Applications, 2004. PG 2004. Proceedings.}, pp.~121--130, IEEE, 2004.

\bibitem{kocon2020head}
M.~Koco{\'n}, ``Head movements in the idle loop animation,'' {\em IADIS Int. J.
  Comput. Sci. Inf. Syst}, vol.~15, no.~2, pp.~137--147, 2020.

\bibitem{asselborn2017}
T.~Asselborn, W.~Johal, and P.~Dillenbourg, ``Keep on moving! exploring
  anthropomorphic effects of motion during idle moments,'' in {\em 2017 26th
  IEEE International Symposium on Robot and Human Interactive Communication
  (RO-MAN)}, pp.~897--902, IEEE, 2017.

\bibitem{vollmer2015alignment}
A.-L. Vollmer, K.~J. Rohlfing, B.~Wrede, and A.~Cangelosi, ``Alignment to the
  actions of a robot,'' {\em International Journal of Social Robotics}, vol.~7,
  no.~2, pp.~241--252, 2015.

\bibitem{vieillard2008}
S.~Vieillard, I.~Peretz, N.~Gosselin, S.~Khalfa, L.~Gagnon, and B.~Bouchard,
  ``Happy, sad, scary and peaceful musical excerpts for research on emotions,''
  {\em Cognition \& Emotion}, vol.~22, no.~4, pp.~720--752, 2008.

\bibitem{russell1980}
J.~A. Russell, ``A circumplex model of affect.,'' {\em Journal of personality
  and social psychology}, vol.~39, no.~6, p.~1161, 1980.

\bibitem{bartneck2009}
C.~Bartneck, D.~Kuli{\'c}, E.~Croft, and S.~Zoghbi, ``Measurement instruments
  for the anthropomorphism, animacy, likeability, perceived intelligence, and
  perceived safety of robots,'' {\em International journal of social robotics},
  vol.~1, no.~1, pp.~71--81, 2009.

\bibitem{barsade2018emotional}
S.~G. Barsade, C.~G. Coutifaris, and J.~Pillemer, ``Emotional contagion in
  organizational life,'' {\em Research in Organizational Behavior}, vol.~38,
  pp.~137--151, 2018.

\bibitem{michalowski2007dancing}
M.~P. Michalowski, S.~Sabanovic, and H.~Kozima, ``A dancing robot for rhythmic
  social interaction,'' in {\em Proceedings of the ACM/IEEE international
  conference on Human-robot interaction}, pp.~89--96, 2007.

\end{thebibliography}

\addtolength{\textheight}{-12cm}

\end{document}